\documentclass[conference]{IEEEtran}
\IEEEoverridecommandlockouts
\usepackage{cite}
\usepackage{amsmath,amssymb,amsfonts}
\usepackage{algorithmic}
\usepackage{graphicx}
\usepackage{textcomp}
\usepackage{xcolor}
\def\BibTeX{{\rm B\kern-.05em{\sc i\kern-.025em b}\kern-.08em
    T\kern-.1667em\lower.7ex\hbox{E}\kern-.125emX}}
\begin{document}

\title{Facial Affect Recognition in the Wild Using Multi-Task Learning Convolutional Network}
\author{\IEEEauthorblockN{Zihang Zhang}
\IEEEauthorblockA{\textit{Nuctech} \\
zhangzihang@nuctech.com}
\and
\IEEEauthorblockN{Jianping Gu}
\IEEEauthorblockA{\textit{Nuctech} \\
gujianping@nuctech.com}
}

\maketitle

\begin{abstract}
This paper presents a neural network based method Multi-Task Affect Net(MTANet) submitted to the Affective Behavior Analysis in-the-Wild Challenge in FG2020. This method is a multi-task network and based on SE-ResNet modules. By utilizing multi-task learning, this network can estimate and recognize three quantified affective models: valence and arousal, action units, and seven basic emotions simultaneously. MTANet achieve Concordance Correlation Coefficient(CCC) rates of 0.28 and 0.34 for valence and arousal, F1-score of 0.427 and 0.32 for AUs detection and categorical emotion classification. 
\end{abstract}


\section{Introduction}
Affective computing attempts to assign computers the capabilities of interpreting and estimation of human affects\cite{b1}. Several channels include visual, auditory and biological signals can develop a system to achieve affective computing. Facial expressions presents the most crucial visual information of human facial affects. \par
Facial Expression Recognition(FER) is the primary topic of facial affect estimation in computer vision field. Recently, there have been numerous studies aim to build an artificial system on FER. However, most current studies pour attention into controlled environments(laboratory environments) because significant variations in the wild increase the difficulty of research. Furthermore, missing in-the-wild dataset limited learning-based methods.\par
Three behavior tasks are annotated to quantify facial expressions: dimensional, Facial Action Coding System(FACS)\cite{b2} and categorical model. In dimensional model, valence and arousal value defined n \cite{b3} indicates an emotion by continuous scale. In FACS model, facial component actions are defined in terms of Action Units(AUs). AUs describe facial muscle movements for further interpreting emotions. In categorical model, seven basic emotions defined by Ekman \emph{et al.}\cite{b4}. Therefore, our multi-task model focus on estimation of above three quantified model.\par
Recently, multi-task learning is a trend of solving complicated problems. According to the annotations, we can solve FER by dividing into three single tasks or designing a multi-task model. Multi-task model assumes that three facial expression quantify methods can work better with each other by sharing similar representation. \par
In this paper, we propose a multi-task deep learning method submitted to the Affective Behavior Analysis in-the-wild Challenge in FG2020 for estimating dimensional model values(valence and arousal), recognizing FACS model(AUs) multi-labels and categorical model(seven basic emotions) label. We report and discuss our result and experiments details.

\section{Related Work}
In recent years, deep learning algorithm and Convolutional Neural Networks(CNN) have become the most popular approach in the field of FER. He \emph{et al.}\cite{b5} who won the ImageNet champion proposed deep residual network(ResNet) to solve the difficulties of training deeper neural network. ResNet stacks residual blocks to build deep network and obtain rich high-level features through learning residual functions. Hu \emph{et al.}\cite{b6} proposed squeeze and excitation network(SENet) to utilize the channel wise attention mechanism on top of deep residual network. Attention module improves residual block with better channel arrangement so that residual block can learn quality features. Woo \emph{et al.}\cite{b7} proposed Convolutional Block Attention Module(CBAM) to develop attention mechanism from channel wise only to spatial wise plus channel wise. CBAM demonstrated that multiple dimensional attentions can cooperate with each other to achieve stronger result.\par
EmotioNet\cite{b8} designed to detect and annotate AUs through recognition of AU and intensity. They derived the relationship between AUs, AU intensity and categorical emotions. Wang \emph{et al.}\cite{b9} proposed Two-level Attention with Two-stage Multi-task(2Att-2Mt) framework which performs convincing result on AffectNet database. They illustrated that applying attention mechanism and multi-task learning helps to improve affect computing in neural network. D.Kollias \emph{et al.}\cite{b10}\cite{b11}\cite{b16}\cite{b19} built a large-scale dataset Aff-Wild and proposed AffWildNet to explain CNN-RNN architectures obtained the best result. Alex \emph{et al.}\cite{b12} proposed to weigh multiple loss functions using homoscedastic uncertainty of each task. They treat the uncertainties as network parameters which can be updated and learned during multi-task learning.

\section{Proposed Methods}
We propose MTANet which is a multi-task learning model based on CNN to accomplish three affect computing tasks: valence and arousal value regression task, action units multi-label classification task and seven basic categorical emotions classification task simultaneously. 

\subsection{Model Structure}
\begin{figure}[htbp]
\centerline{\includegraphics[scale=0.8]{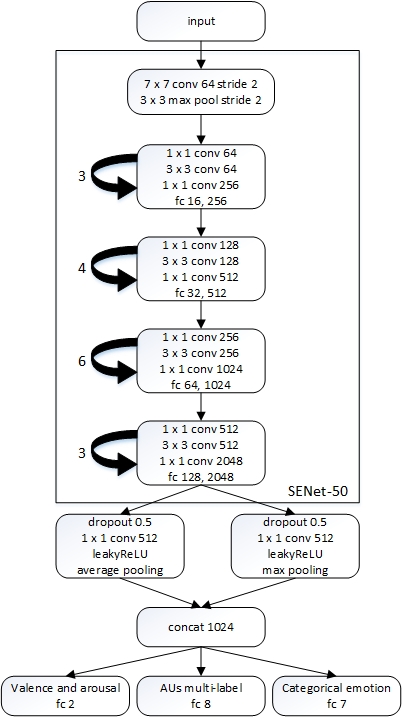}}
\caption{Network architecture of MTANet. The number represents output channel size.}
\label{fig}
\end{figure}
As shown in Fig.~\ref{fig}, we employ SENet-50 as backbone to extract 2048-d features from input 112*112 aligned face images. An average pooling and max pooling layer with dropout and ReLU activation function followed by backbone tends to refine stronger high-level features. This encoder eventually produces a 1024-d feature map after concatenation. We design hard parameter sharing multi-task learning encoder to develop a powerful encoder by taking advantages of numerous annotated data. In decoder model, we simply apply fully connected layer for each task. Dimensional model layer's output size is 2 because of two regression values. AUs model layer's output size is 8 for eight individual AU labels. Categorical model layer's output size equals to the number of annotated basic emotions.

\subsection{Loss Function}
We design for every individual task. We employ categorical cross entropy loss for basic emotion classification, binary cross entropy for AUs multi-label classification and Concordance Correlation Coefficient(CCC) for valence and arousal regression. Total loss is the sum of all individual loss.
\begin{equation}
L_{expr} = -x[class] + log\sum^6_{i=0}e^{x_i}\label{eq1}
\end{equation}
\begin{equation}
L_{au} = -\sum^7_{i=0}[y_i*log\sigma(x_i) + (1-y_i)*log(1-\sigma(x_i))]\label{eq2}
\end{equation}
\begin{equation}
\rho_{ccc} = \dfrac{2S_{xy}}{[s^2_x +s^2_y +(\overline x-\overline y)^2]}\label{eq3}
\end{equation}
\begin{equation}
L_{ccc} = 1 - \dfrac{1}{2}[\rho_{a}+\rho_{v}]\label{eq4}
\end{equation}
Equation \eqref{eq1} is the categorical cross entropy loss for categorical emotion classification where class indicates the index in the range of [0, 6]. AUs multi-label classification task employs equation \eqref{eq2} as the loss function where $$y_i\in[0,1]$$ states the AUs ground truth label. Equation \eqref{eq3} is the equation of computing valence and arousal CCC value and equation \eqref{eq4} is objective function of dimensional emotion model.
\begin{equation}
\begin{split}
L_{weighted} = &\dfrac{1}{2\sigma^2_{va}}L_{va} + \dfrac{1}{\sigma^2_{au}}L_{au} + \dfrac{1}{\sigma^2_{expr}}L_{expr}\\&+ log\sigma_{va} + log\sigma_{au} + log\sigma_{expr} \label{eq5}
\end{split}
\end{equation}
We utilize two types of total loss function: simply sum up all individual task loss and weighted total loss by implementing the equation \eqref{eq5} in \cite{b12}. We firstly train our model with simply sum method and then fine-tuned the model with weighted total loss function.

\section{Database \& Result}
In this section, we introduce Aff-Wild2 database provided for the Affect-in-the-Wild challenge. Next, we address our experiment details and results to demonstrate that MTANet is able to  results.
\subsection{Aff-Wild2 database}
Aff-Wild2\cite{b17}\cite{b18} database is an extension upon Aff-Wild database which a large-scale, sufficiently annotated and dynamic in the wild database. There are several benefits of Aff-Wild2 database:
\begin{itemize}
\item It is a large-scale in-the-wild database contains over five hundred videos and large number of objects.
\item It is a dynamic audiovisual database providing the conditions where researchers can build CNN-RNN multi-modal and multi-task model.
\item It includes all main behavior tasks: dimensional, categorical, and FACS and all labels are manually annotated.
\end{itemize}

\subsection{Experiment result}
\begin{table}[htbp]
\caption{Results on validation set}
\begin{center}
\begin{tabular}{|c|c|c|c|c|}
\hline
\textbf{Backbone}&\multicolumn{4}{|c|}{\textbf{Validation Set}} \\
\cline{2-5} 
\textbf{Structure} & \textbf{Valence}& \textbf{Arousal}& \textbf{AUs}& \textbf{Expression} \\
\hline
Baseline& 0.14 & 0.24& 0.31 & 0.36 \\
ResNext-50& 0.22 & 0.31& 0.403 & 0.30 \\
ResNext+CBAM& 0.27 & 0.34& 0.397 & 0.31  \\
SENet-50& 0.28 & 0.34& 0.427 & 0.32  \\
\hline
\end{tabular}
\label{tab1}
\end{center}
\end{table}
Our proposed method is implemented using a combination of PyTorch\cite{b14} and Scikit-learn\cite{b15} toolbox on NVIDIA Tesla V100 and RTX2070S. We used Adam optimizer with learning rate of 0.001 and weight decay of 0.001 in training process and stochastic gradient descent with learning rate of 0.0001 and weight decay of 0.005 in fine tune process. We utilized SENet-50 pretrained model on VGGFace2.\par
We experiment three backbones include: ResNext-50, ResNext-50 with CBAM module and SENet-50 to figure out the best hard parameter share structure. As shown in Table.~\ref{tab1}, SENet-50 achieve the best performance between four models. ResNext with CBAM obtains similar result with SENet therefore illustrating the effectiveness of attention mechanism. Compare to baseline result\cite{b13}, SENet-50 performs well on valence and arousal regression and AU multi-label classification task, but not on categorical emotion classification task. We also attempt to find out a better combination of linear layers by stacking deeper linear layers. However, complicated linear layer design increases parameters and cast no influence on result. 

\section{Conclusion}
In this paper, we proposed MTANet submitted to the Affect-in-the-Wild Challenge. Aff-Wild2 database is an excellent dataset provides the conditions where multi-modal and multi-task learning network maximize their advantages. We addressed that multi-task learning and attention mechanism can help deep neural network to obtain remarkable result in affect computing. Multi-task learning improves every individual task performance through efficient and stronger hard parameter share encoder. We demonstrated that MTANet achieves better result than the baseline result.

\section*{Acknowledgment}
We gratefully acknowledge the support from D.Kollias with his discussion and patience.



\end{document}